\documentclass[10pt,twocolumn,letterpaper]{article}
\usepackage{graphicx}
\usepackage{multirow}
\usepackage{cvpr}              
\usepackage{tabularx,colortbl}

\definecolor{cvprblue}{rgb}{0.21,0.49,0.74}
\usepackage[pagebackref,breaklinks,colorlinks,allcolors=cvprblue]{hyperref}
\usepackage{multirow}
\def\paperID{*****} 
\def\confName{CVPR}
\def\confYear{2025}

\title{LLMPi: Optimizing LLMs for High-Throughput on Raspberry Pi}

\author{Mahsa Ardakani, Jinendra Malekar, Ramtin Zand\\
Department of Computer Science and Engineering, University of South Carolina\\
{\tt\small mahsam@email.sc.edu, jmalekar@email.sc.edu, ramtin@cse.sc.edu}}

\begin{document}
\maketitle
\begin{abstract}
Deploying Large Language Models (LLMs) on resource-constrained edge devices like the Raspberry Pi presents challenges in computational efficiency, power consumption, and response latency. This paper explores quantization-based optimization techniques to enable high-throughput, energy-efficient execution of LLMs on low-power embedded systems. Our approach leverages k-quantization, a Post-Training Quantization (PTQ) method designed for different bit-widths, enabling efficient 2-bit, 4-bit, 6-bit, and 8-bit weight quantization. Additionally, we employ ternary quantization using Quantization-Aware Training (QAT) for BitNet models, allowing for more effective adaptation to lower-bit representations while preserving accuracy.

Our findings highlight the potential of quantized LLMs for real-time conversational AI on edge devices, paving the way for low-power, high-efficiency AI deployment in mobile and embedded applications. This study demonstrates that aggressive quantization strategies can significantly reduce energy consumption while maintaining inference quality, making LLMs practical for resource-limited environments.
\end{abstract}    
\vspace{-8mm}
\section{Introduction}
\label{sec:intro}
Conversational AI has revolutionized human-computer interaction, enabling real-time, intelligent dialogue in domains such as customer service, healthcare, and education \cite{sanchez2024automating}. One key application is social robotics, where AI-powered systems enable personalized, real-time interactions in education, elderly care, and assistive technologies \cite{breazeal2016social, belpaeme2018social}. These systems leverage natural language processing (NLP) and speech technologies to improve accessibility and engagement. 

The rise of large language models (LLMs) has significantly advanced NLP, enabling more sophisticated natural language understanding and generation. Models like GPT\cite{brown2020language}, BERT \cite{devlin2019bert}, and LLaMA \cite{touvron2023llama} leverage deep transformer architectures and large-scale training data to produce highly contextual and coherent responses, improving conversational AI across various domains. 

However, real-world deployment of conversational AI systems requires low-latency, on-device AI inference, as cloud-based solutions introduce privacy risks, response delays, and reliance on network connectivity. Edge computing mitigates these challenges by enabling faster, decentralized processing, improving privacy, efficiency, and responsiveness \cite{varghese2016challenges,reidy2023efficient, khan2019edge,mohammadi2023facial,reidy2023work,smith2023realtime,mohammadi2024edge}. 

Among edge devices, Raspberry Pi \cite{ghael2020review} is widely adopted for its affordability, portability, and energy efficiency, making it a popular choice in healthcare, automation, and robotics. However, deploying LLMs on Raspberry Pi presents computational constraints, necessitating efficient optimization strategies. 
A real-time conversational AI system \cite{kulkarni2019conversational} typically consists of three core components: (a) A voice-to-text module for processing user input for the LLM, (b) An LLM that generates responses dynamically, and (c) A text-to-speech generator optimized for real-time processing. Among these, LLM inference is the most computationally intensive, traditionally requiring high-performance GPUs or TPUs. To enable real-time AI on resource-limited hardware, model compression techniques such as quantization are essential.

This paper explores quantization-based optimization for efficient LLM deployment on Raspberry Pi. We apply post-training quantization (PTQ) across multiple bitwidths including W8A8 (8-bit weight, 8-bit activation), W6A8, W4A8, and W2A8. We also explore the efficiency of BitNet models \cite{wang2023bitnet,ma2024era,malekar2024matmul}, which are inherently optimized for extreme low-bit representations using quantization-aware training (QAT). Specifically, in this paper, we study BitNet models with ternary (-1, 0, 1) weights, referred to as Q1.58, and 8-bit activations. We evaluate quantized models against 16-bit floating-point (FP16) baselines, analyzing inference speed, energy efficiency, and response quality. Through a systematic evaluation, this study demonstrates that extreme quantization strategies significantly reduce computational overhead while preserving inference quality, making real-time LLM deployment feasible for social robotics, mobile assistants, and embedded AI applications.

\vspace{-2mm}
\section{Related Work}

\subsection{LLM Model Compression}


Model compression can be achieved through various techniques, including pruning (structured and unstructured), knowledge distillation (white box and black box), quantization (QAT and PTQ), and low-rank factorization. Recently, there has been significant progress in combining these strategies for lossless model compression. One notable approach is LLM-Pruner \cite{NEURIPS2023_44956951}, which integrates structured pruning with low-rank factorization. LLM-Pruner employs gradient information to prune coupled structures in a structured manner and recovers the efficiency of the pruned model through fine-tuning techniques such as Low-Rank Adaptation (LoRA) \cite{hulora}.

QLoRA \cite{NEURIPS2023_1feb8787} is a method primarily used for efficient fine-tuning rather than model compression. It combines LoRA and custom quantization techniques, such as block-wise quantization and double quantization, to enable loading the model in 4-bits. This approach highlights efficient compression methods that can be employed for model compression. Among other effective methods are MiniLLM \cite{gu2024minillm} and AWQ \cite{lin2024awq}. MiniLLM utilizes knowledge distillation from larger white-box LLMs to smaller models, employing reverse Kullback-Leibler divergence (KLD) instead of standard KLD, which is more suitable for generative LLMs. AWQ is a low-bit weight-only quantization method that suggests not all weights in an LLM are crucial. By protecting only 1\% of the salient weights identified through activation distribution rather than weight distribution, AWQ significantly reduces quantization errors. Combined with the TinyChatEngine inference framework, this approach resulted in a 3$\times$ speedup compared to FP16 LLMs on both desktop and mobile GPUs. 


\subsection{Transformer Quantization}


Transformer quantization can be classified into two main types: weight-only quantization and weight-and-activation quantization. The former primarily reduces memory usage, while the latter also improves computational efficiency. 

SmoothQuant \cite{xiao2023smoothquant} enables both weight and activation quantization by mathematically transforming the challenge of activation quantization. This approach achieves a 1.5$\times$ speedup and a 2$\times$ reduction in memory usage for LLMs with minimal performance loss, supporting W8A8 (8-bit weight, 8-bit activation) quantization.
Omniquant \cite{shao2023omniquant} extends support to a wider range of weight-activation quantization configurations, including W4A4, W4A16, W3A16, and W2A16. It accomplishes this through learnable weight clipping, which optimizes the clipping threshold, and learnable equivalent transformation, which shifts activation outliers to weights for better quantization stability.
LLM.int8() \cite{dettmers2022gpt3} loads models in 8-bit format, performing 99.9\% of computations in 8-bit while handling rare outliers separately. It employs vector-wise quantization with distinct normalization constants to maintain model accuracy.
GPTQ \cite{frantar2023optq} focuses exclusively on weight quantization, achieving 3-bit and 4-bit quantization with speedups of 3.24$\times$ and 4.53$\times$ on A6000 and A100 GPUs, respectively.
QuIP\# \cite{tseng2024quipbetterllmquantization} is a weight-only quantization technique that delivers state-of-the-art performance in sub-4-bit quantization. It leverages a randomized Hadamard transform combined with vector quantization, followed by fine-tuning to improve model fidelity. All the methods mentioned above fall under post-training quantization (PTQ).

\subsection{1-bit LLMs}


Recently, following the rise of LLMs and advancements in model performance, the BitNet \cite{wang2023bitnet} introduced a 1-bit transformer quantization technique for LLMs. This method replaces the conventional `nn.Linear` layer in PyTorch with a new BitLiner layer, where weights are restricted to either 1 or -1, and activations are represented in 8-bit precision. Despite this quantization, other components like self-attention remain in 8-bit format. BitNet's design suggests that it can scale to even larger transformer models, following scaling laws similar to those used for full-precision transformers. A variant of BitNet, known as BitNet 1.58 \cite{ma2024era}, employs ternary weights (-1, 0, 1) and achieves perplexity and end-task performance comparable to full-precision transformers (FP16 or BF16). The 1-bit LLMs transform MatMul operations into addition operations, due to the 1-bit nature of the weights, except for layers like attention, which need to remain in high precision to maintain performance. 

\section{Underlying Operations of LLMs}

\begin{figure*}[t]
\centering
\includegraphics[width=0.98\textwidth]{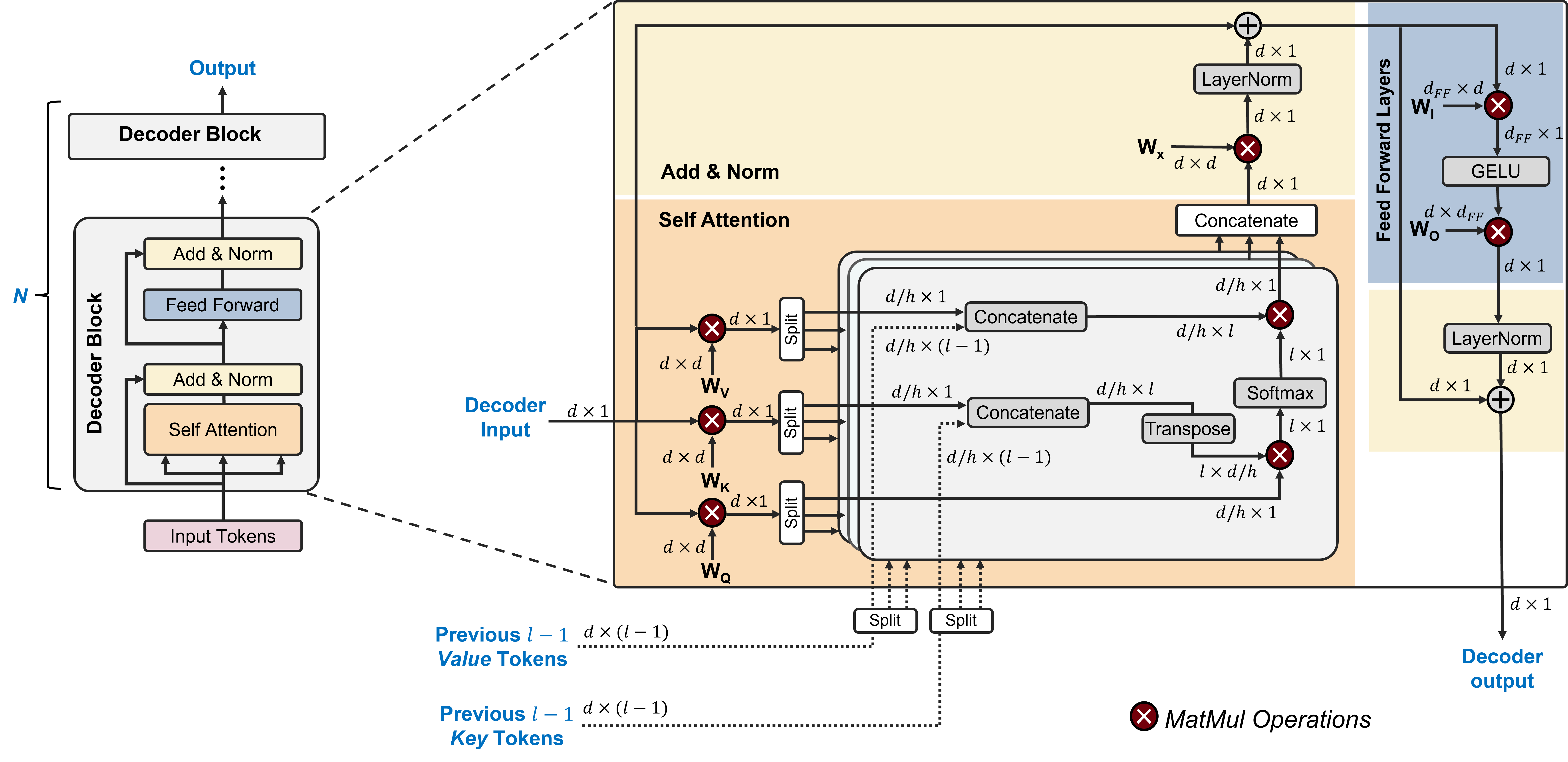}
\caption{The typical architecture of decoder-only LLM models and its underlying operations.} 
\label{fig:decoder}
\end{figure*}


The overall architecture of the generative LLMs is shown in Figure \ref{fig:decoder}, which includes $N$ decoder blocks, each consisting of self-attention and feedforward layers followed by add and normalization operations \cite{vaswani2017attention}. The core of the LLM is the self-attention mechanism with multiple heads ($h$). For an attention head, the computation begins with three linear projections of the token vector to form the Key ($K$), Query ($Q$), and Value ($V$) vector sequences:
\vspace{-1mm}
\begin{equation}
    Q = W_Q*I, \hspace{9pt} 
    K = W_K*I, \hspace{9pt}
    V = W_V*I
\end{equation}

\noindent where $I$ in the input token vector and $W_Q$, $W_K$, and $W_V$ are trainable weight matrices with $d \times d$ dimensions where $d$ is the embedding dimension. The operator $*$ denotes the MatMul operation.

The generated $K$, $Q$, and $V$ vectors are then divided into $h$ vectors with reduced dimensionality of $d/h$ where $h$ is the number of attention heads. At this stage, the value and key vectors are concatenated with the previous $l-1$ value and key tokens that are cached from previous token generation iterations to form a $d/h \times l$ matrix in each head, where $l$ is the sequence length. Next, the attention scores are computed using the scaled dot-product of the queries and keys multiplied with the generated value matrix \cite{vaswani2017attention}. 

\vspace{-1mm}
\begin{equation}
    \text{Attention} (Q, K, V) = softmax (\frac{Q*K^T}{\sqrt{d}}) * V
\end{equation}

\vspace{-1mm}
Subsequently, the output of the attention heads are concatenated and linearly transformed as follows:
\vspace{-1mm}

\begin{equation}
    \text{MultiHead} (Q,K,V) = \text{Concat} (head_1, ..., head_h) * W_X
\end{equation}

\noindent where $head_i$ = Attention $(Q_i, K_i, V_i)$, and $W_X$ is a $d \times d$ matrix with trainable elements.

Finally, the attention output is followed by a feed-forward network (FFN) that involves two linear and one nonlinear transformations. As shown in Figure \ref{fig:decoder}, Gaussian error linear unit (GELU) \cite{gelu} activation function is frequently utilized in LLMs to introduce nonlinearity into FFN. Therefore, the FFN computation can be described as follows, which involves two more MatMul operations: 




\begin{equation}
    FFN (x) = GELU (x*W_I + b_I)*W_O + b_O
\end{equation}

\noindent where $W_I$ and $W_O$ are trainable matrices with $d_{FF} \times d$ and $d \times d_{FF}$ dimensions, respectively, where $d_{FF}$ is the feed-forward layer size. Also, the $x$ is \text{MultiHead} $(Q,K,V)$. Both the self-attention and FFN layers are followed by layer normalization which can also introduce some nonlinearity in the process of division by the standard deviation.

\begin{table}[]
\centering
\begin{tabular}{lcc}
\hline
Block                                                                           & Description     & Dimension                                                               \\ \hline
\multirow{4}{*}{\begin{tabular}[c]{@{}l@{}}Attention\\Projections\end{tabular}} & $W_Q$            & \multirow{4}{*}{$(d\times d)* (d\times 1)$} \\
                                                                                & $W_K$            &                                                                         \\
                                                                                & $W_V$            &                                                                         \\
                                                                                & $W_X$            &                                                                         \\ \hline
\multirow{2}{*}{\begin{tabular}[c]{@{}l@{}}Attention\\ Head\end{tabular}}       & $Q*K$             & $(l\times d/h) * (d/h\times 1)$            \\
                                                                                & $V*Score$         & $(d/h\times l) * (l\times 1)$               \\ \hline
\multirow{2}{*}{FFN}                                                            & Intermediate FF & $(d_{FF} \times d) * (d\times 1)$            \\
                                                                                & Output FF       & $(d\times d_{FF}) * (d_{FF}\times 1)$         \\ \hline
\end{tabular}
\caption{MatMuls in LLMs. The parameters $d$, $h$, $l$, and $d_{FF}$ denote embedding dimension, number of attention heads, sequence length, and the size of the feed-forward layers, respectively.  }
\label{tab:dimension}
\end{table}

In summary, the computation of decoder blocks in LLMs involves a combination of nonlinear operations (LayerNorm, softmax, GELU) and linear matrix multiplications (MatMul). 
In decoder-only LLMs, the MatMuls are matrix-vector multiplications since processing is done iteratively during inference, caching keys and values from previous iterations, as illustrated in Figure~\ref{fig:decoder}. Table~\ref{tab:dimension} provides the dimensions of each MatMul operation.

Prior research \cite{kim2023full} has demonstrated that specialized hardware for nonlinear operations can minimize their computational overhead to the point of insignificance compared to MatMul operations. As a result, optimizing MatMul operations plays a crucial role in accelerating LLM inference. Quantization techniques, including PTQ and QAT, offer the potential to substantially reduce both the computational and memory demands of MatMuls.

\section{Approach: K-Quantization}

To reduce the memory footprint of LLMs and enhance their throughput in conversational AI systems, we apply four levels of quantization: W8A8, W6A8, W4A8, and W2A8 using PTQ. Additionally, we employ ternary-weight models with 8-bit activations through quantization-aware trained Bitnet models. These quantized models are evaluated against their full precision (FP16) baselines. 

In this work, we use the k-quantization approach, a PTQ method, to implement 8-bit, 6-bit, 4-bit, and 2-bit quantization. The k-quantization method may differ based on the bit quantization level applied, but its core process is described as follows.

First, the weights of the layer are divided into “super” blocks, each containing multiple “sub” blocks. From these blocks, the scale factor (\(s\)) and alpha (\(\alpha\)) are derived. (\(\alpha\)) is the largest value within the block, and (\(s\)) is defined as: 
\[
(s)  = \frac{2^b - 1}{\alpha}
\]

\noindent where $b$ is the bit width of the quantized value.

Next, to quantize a given “sub” block, the $absmax$ quantization method is applied. This involves multiplying the weights by the scale factor (\(s\)):  

\[
w_{\text{quantized}} = w \cdot s
\]

The scale factor for a ``sub" block is calculated using the information from that block, but it is quantized according to the scale factor of its corresponding ``super" block. Specifically, the block-wise quantization method uses the scale factor from the “super” block (\(s_{\text{super}}\)) to quantize the scale factor of the “sub” block (\(s_{\text{sub}}\)) as expressed below:

\[
s_{\text{sub, quantized}} = s_{\text{sub}} \cdot s_{\text{super}}
\]

 The precision of scale factors may differ, with the “super” block generally having a higher precision than the “sub” block. In this work, for W2A8 and W4A8, we apply the k-quantization method variants \texttt{Q2\_K} and \texttt{Q4\_K}. The \texttt{Q2\_K} variant uses 2 bits per weight, with super-blocks containing 16 blocks, each having 16 weights. Similarly, the \texttt{Q4\_K} variant uses 4 bits per weight in super-blocks containing 8 blocks, each with 32 weights.








The \texttt{Q1.58} is a special case of quantization, as it is the only quantization type that utilizes ternary weights (\{-1, 0, 1\}). A model with ternary weights requires quantization-aware training, as accuracy would decrease significantly without it \cite{ma2024era}. To ensure consistency with our experiments and enable comparison with other models quantized using PTQ, we use off-the-shelf \texttt{BitNet1.58} models \cite{ma2024era} featuring ternary weights and 8-bit activations. BitNet performs quantization within the transformer architecture by replacing linear layers with BitLinear layers, rather than directly quantizing the tensors. It employs \textit{absmax} quantization to quantize the weights in these layers.


\begin{table}
\centering
\begin{tabular}{l| c c c c}
\hline
Models       & $d$    & $h$  & $d_{FF}$   & $l$                  \\ \hline
Gemma 2      & 2048 & 8  & 16384 & 8192               \\ \hline
Llama3.2-1B  & 2048 & 32 & 8192  & 131072 (8192)      \\ 
Llama3.2-3B  & 3072 & 24 & 8192  & 131072 (8192)      \\ 
Llama3-8B  & 4096 & 32 & 14336 & 8192               \\ \hline
Phi3         & 3072 & 32 & 8192  & 4096               \\ \hline
Bitnet0.7B   & 1536 & 24 & 4096  & 2048               \\ 
Bitnet\_8B   & 4096 & 32 & 14336 & 8192               \\ \hline
\end{tabular}
\caption{Model specifications}
\label{tab:model_specs}
\end{table}

\section{Experimental Setup}

\textbf{Performance Metrics:} To evaluate the impact of quantization techniques, we 
use key performance metrics, including latency, throughput, accuracy, and energy efficiency. Specifically, we measure tokens per second (TPS) to assess inference speed, tokens per joule (TPJ) to quantify energy efficiency, words per battery life (W/BL) to estimate real-world feasibility on battery-powered devices, and the NUBIA Score \cite{kane2020nubia} to evaluate semantic coherence and response quality in generated text. These energy efficiency metrics (TPJ and W/BL) are particularly important for assessing LLM deployment in resource-constrained environments, balancing power consumption, model performance, and real-time usability on edge-based AI applications.

\vspace{2mm}
\noindent \textbf{Dataset:} For benchmarking, we employ the Stanford Question Answering Dataset (SQuAD) v2 \cite{rajpurkar2016squad}, specifically its validation set, to extract relevant contexts and questions for testing. SQuAD is a widely recognized benchmark for reading comprehension tasks, providing a structured evaluation framework for assessing model-generated responses in terms of accuracy and contextual relevance. 

\vspace{2mm}
\noindent \textbf{Models:} The evaluation includes multiple LLM architectures, including Llama-1B, Llama-3B, Llama-8B \cite{dubey2024llama}, Phi-3B \cite{abdin2024phi}, and Gemma-2B \cite{team2024gemma}, all of which have fewer than 8 billion parameters. The hyperparameters of these models are listed in Table \ref{tab:model_specs}. These models were chosen for their suitability in resource-constrained edge deployments, specifically on the Raspberry Pi 5 with 8GB of memory. To accommodate memory limitations and enable efficient processing, an M.2 HAT expansion module is integrated into the testing setup.

We evaluate models across multiple precision levels, from full precision (FP16) to lower-bit configurations, including 8-bit (Q8), 6-bit (Q6), 4-bit (Q4), and 2-bit (Q2). Additionally, we incorporate BitNet models—specifically BitNet-0.7B and BitNet-8B—which inherently use ternary weights ({-1, 0, 1}) with 8-bit activations, following a Q1.58 quantization scheme \cite{ma2024era}. 


\begin{table}[]
\small 
\centering
\begin{tabular}{l c c c}
\hline
 &  & \multicolumn{2}{c}{Power (W)} \\
\multirow{-2}{*}{Models} & \multirow{-2}{*}{Latency (ms)} & Dynamic & Total \\\hline
Llama1B\_FP   & 262.81  & 3.13  & 6.45 \\
Llama1B\_Q2   & 67.68   & 2.85  & 6.16 \\
Llama1B\_Q4   & 85.63   & 2.77  & 6.09 \\
Llama1B\_Q6   & 106.92  & 3.01  & 6.33 \\
Llama1B\_Q8   & 134.58  & 3.04  & 6.35 \\\hline
Gemma2B\_FP   & 512.03  & 3.65  & 6.97 \\
Gemma2\_Q2    & 149.81  & 3.17  & 6.48 \\
Gemma2\_Q4    & 178.38  & 3.05  & 6.37 \\
Gemma2\_Q6    & 230.73  & 3.16  & 6.47 \\
Gemma2\_Q8    & 278.39  & 3.18  & 6.50 \\\hline
Llama3B\_FP   & 680.27  & 2.90  & 6.21 \\
Llama3B\_Q2   & 196.38  & 3.44  & 6.76 \\
Llama3B\_Q4   & 201.53  & 3.27  & 6.58 \\
Llama3B\_Q6   & 294.20  & 3.14  & 6.46 \\
Llama3B\_Q8   & 363.63  & 3.24  & 6.55 \\\hline
Phi3B\_FP   & 10309.28  & 3.23  & 6.55 \\
Phi3B\_Q2   & 237.92  & 3.39  & 6.71 \\
Phi3B\_Q4   & 242.48   & 3.24  & 6.55 \\
Phi3B\_Q6   & 365.76  & 3.38  & 6.70 \\
Phi3B\_Q8   & 444.04  & 3.18  & 6.50 \\\hline
Llama8B\_FP   & 33333.33  & 2.76  & 6.08 \\
Llama8B\_Q2   & 471.47  & 3.74  & 7.06 \\
Llama8B\_Q4   & 467.94  & 3.35  & 6.66 \\
Llama8B\_Q6   & 726.21  & 3.36  & 6.67 \\
Llama8B\_Q8   & 8695.65  & 3.66  & 6.97 \\\hline
BitNet\_Q1   & 52.00  & 4.12  & 7.43 \\
BitNet\_Q1.58   & 510.20  & 3.54  & 6.86 \\\hline

\end{tabular}
\caption{Performance comparison of different models.}
\label{tab:performance_comparison}
\end{table}

\vspace{2mm}
\noindent \textbf{Experimental Measurements:} Table~\ref{tab:performance_comparison} presents the data collected to measure dynamic power, latency, and total power across different model configurations. To quantify energy consumption, we measure static power, dynamic power, and total power using a USB 3.0 multimeter, \textit{i.e.}, MakerHawk UM34C, connected to the Raspberry Pi. Power dissipation is recorded in two phases:

\begin{itemize}
    \item Static Power: Measured over a three-minute idle period without model execution.
    \item Total Power: Recorded during three minutes of inference, capturing real-time power usage.
\end{itemize}
The dynamic power consumption is calculated as the difference between the total power and static power.

To ensure a fair benchmarking process, all inference evaluations use a standardized prompt with a context length of 4096 tokens, ensuring consistent TPS measurements across models. However, for NUBIA Score evaluation, the context length is extended to 8192 tokens to enable a comprehensive assessment of semantic coherence and response consistency over longer sequences.


\section{Results}
\begin{figure}[]
  \centering
    \includegraphics[width=1.0\linewidth]{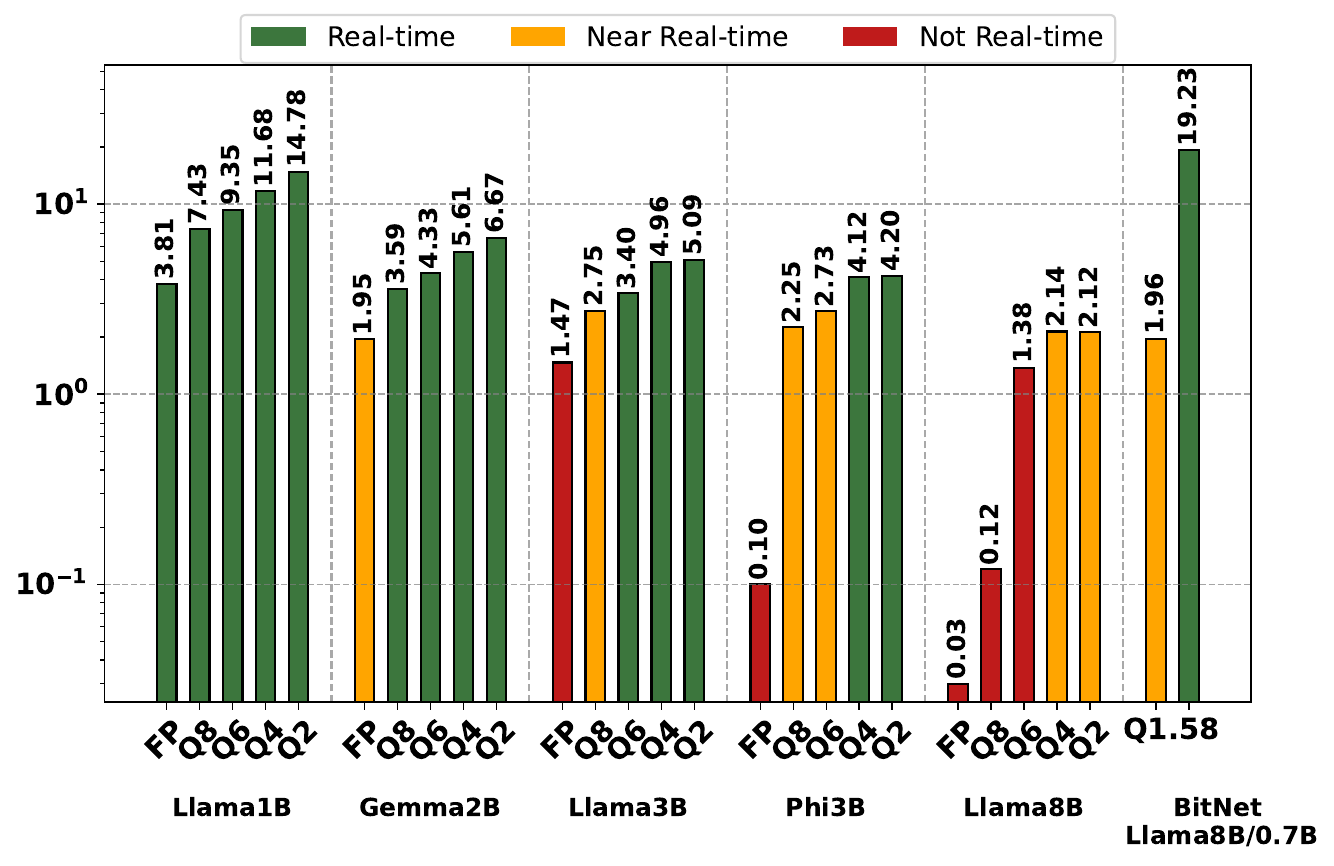}
      \caption{TPS across different LLMs and precision levels.}

    \label{fig:TPS}
\end{figure}

\subsection{Throughput}

Figure~\ref{fig:TPS} presents the TPS across various model configurations, with color-coded bars indicating real-time feasibility. We define a minimum TPS of 3 as the threshold for seamless human-AI conversation. This is based on an estimated speaking rate of 120 words per minute \cite{rayner2009language} and a conservative assumption of 1.5 tokens per word \cite{lambruschini2023reducing}, requiring an end-to-end system throughput of at least 3 TPS for real-time interaction.



Models achieving TPS above 3 ensure smooth token generation with minimal latency, while those in the 2 to 3 range may experience slight delays but remain functional. In contrast, models below 1.5 TPS exhibit significant latency, making them unsuitable for real-time applications. The BitNet models are positioned from left to right in the figure as \texttt{BitNet\_8B} followed by \texttt{bitnet0.7B}. These results highlight the trade-offs between computational speed and precision reduction, demonstrating the impact of quantization on inference latency and model efficiency in resource-constrained environments.

A general trend emerges across all models, demonstrating that higher levels of quantization consistently lead to improved TPS. This improvement underscores the trade-off between precision and computational speed, particularly in resource-constrained environments. While some smaller models, such as Llama1B, achieve real-time performance in FP16, others, like Gemma2B, Llama3B, and Phi3B, require quantization to reach real-time speeds. For instance, Llama1B improves from 3.81 TPS in FP16 to 14.78 TPS in Q2, representing a 3.88$\times$ acceleration. Similarly, Gemma2B transitions from near real-time performance at 1.95 TPS in FP16 to a real-time range of 6.67 TPS in Q2. These results highlight the suitability of smaller models for low-latency applications and emphasize the benefits of quantization for enhancing computational efficiency.

However, the impact of quantization becomes even more pronounced for larger models, such as Llama8B. While these models benefit significantly from aggressive quantization, achieving real-time feasibility remains a challenge. For example, Llama8B starts at just 0.03 TPS in FP16 but accelerates to 2.14 TPS in Q4—an impressive 71$\times$ speedup. Although these values still fall short of the real-time threshold, they represent significant progress toward making large-scale models viable for edge deployment. Moderate quantization levels (Q8 and Q6) push Llama8B into the near real-time range, reaching 0.12 and 1.38 TPS, respectively. However, achieving real-time performance for such large-scale models may require further optimizations, such as mixed precision techniques or advanced compression strategies.

Among these improvements from quantization, \texttt{BitNet0.7B} stands out as the top performer, achieving the highest TPS at 19.23—far surpassing other models. Unlike traditional transformer-based LLMs, BitNet replaces conventional matrix multiplications with optimized bitwise operations using BitLinear layers. This architectural innovation dramatically enhances inference speed, making BitNet a compelling choice for real-time applications.


%

\begin{figure}[]
    \centering
    \includegraphics[width=1.0\linewidth]{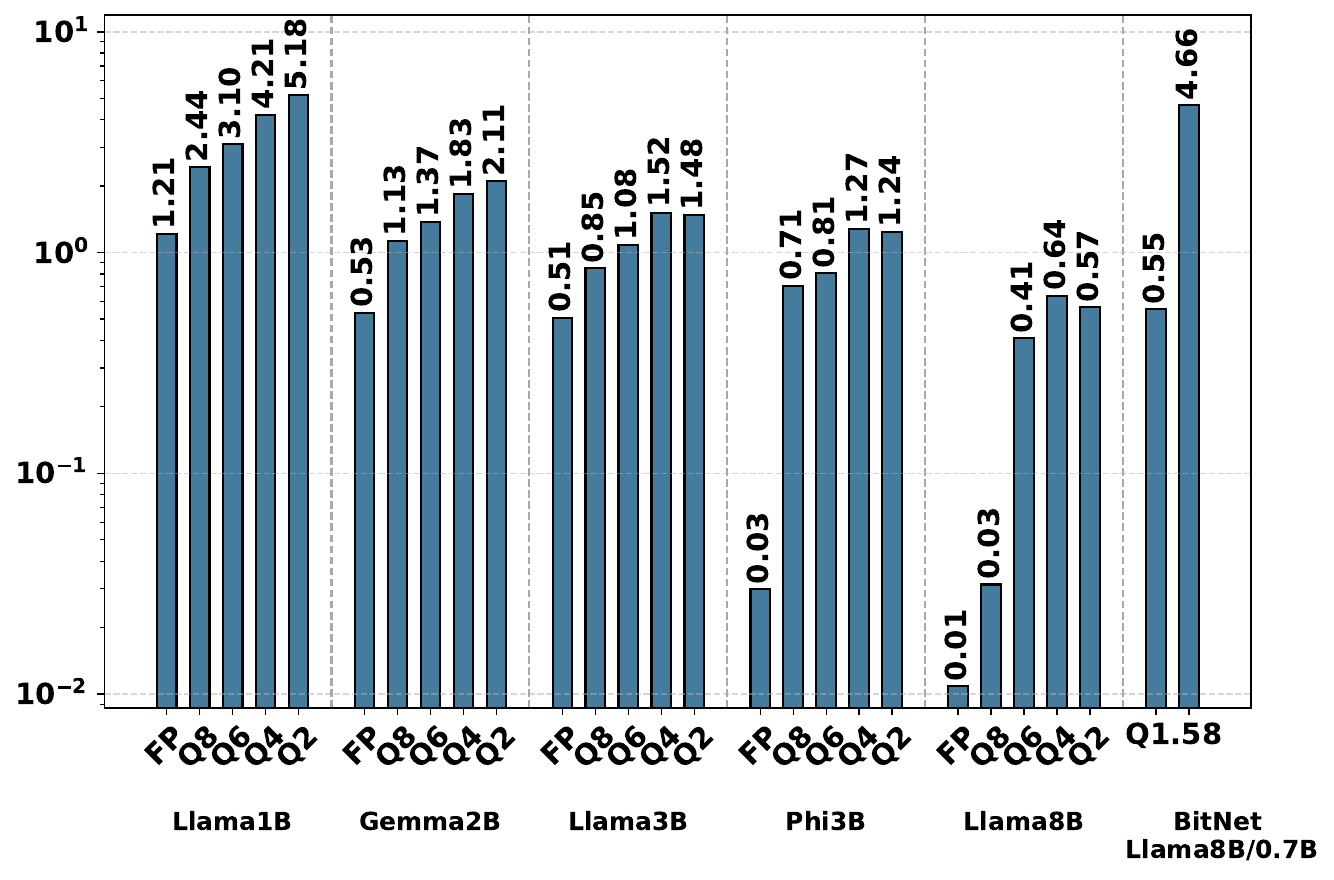}
    \caption{Energy efficiency of LLMs measured in TPJ.}
    \label{fig:TPJ}
\end{figure}

\subsection{Tokens per Joule}
Figure \ref{fig:TPJ} presents the energy efficiency of various LLM configurations, measured in TPJ. 
Higher TPJ values indicate greater energy efficiency, meaning the model generates more tokens per unit of energy consumed, which is essential for real-time applications under strict power budgets. 

Quantization significantly impacts energy efficiency, with lower-bit models consistently demonstrating higher TPJ. Full-precision (FP16) models have the lowest TPJ, emphasizing their high energy consumption. Among all models, Llama1B\_Q2 achieves the highest TPJ at 5.18, representing a 4.3$\times$ energy efficiency improvement over its FP16 counterpart (1.21 TPJ). This makes Llama1B\_Q2 the most energy-efficient model in this study. Interestingly, Llama1B\_Q2 is also 1.11$\times$ more energy-efficient than BitNet0.7B\_Q1.58 (4.66 TPJ), suggesting that small models with aggressive PTQ can slightly outperform BitNet’s QAT approach. Nonetheless, BitNet0.7B\_Q1.58 remains highly competitive, surpassing the Q2 configurations of all other models, demonstrating that specialized bitwise operations can achieve energy efficiency on par with conventional quantized transformers. Gemma2B follows a similar trend, improving from 0.53 TPJ in FP16 to 2.11 TPJ in Q2, reinforcing that lower-bit quantization significantly enhances energy efficiency.

\begin{figure}[]
    \centering
    \includegraphics[width=1.0\linewidth]{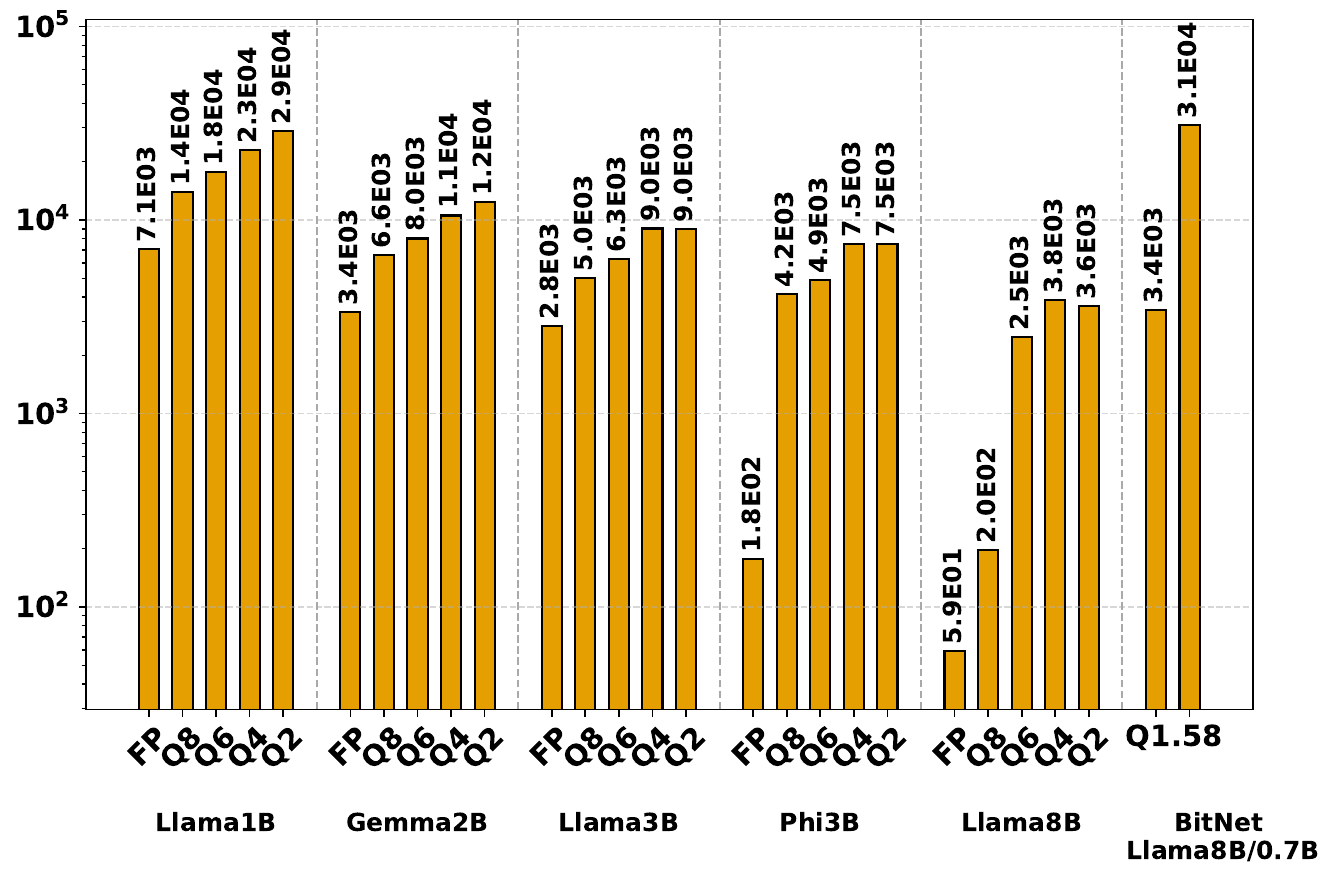}
    \caption{Words per Battery Life (W/BL) for different LLMs and precision levels.}
    \label{fig:WBL}
\end{figure}

\subsection{Words per Battery Life}
Figure \ref{fig:WBL} shows a comparative analysis of Words per Battery Life (W/BL), which quantifies the number of words a model can generate within a single battery cycle. This estimation is based on a typical edge device battery capacity of 5 watt-hours (18,000 Joules) and an assumed average of 1.5 tokens per word. W/BL is used to assess the feasibility of deploying LLMs in power-constrained environments, such as mobile devices and edge computing platforms, where maximizing word generation per battery cycle is crucial for real-world usability. 

The results indicate that lower-bit quantization significantly enhances W/BL, leading to greater energy efficiency and reduced power consumption per generated word. Full-precision (FP16) models exhibit the lowest W/BL values, reinforcing their high energy demands. As bit-width decreases, quantized models can generate substantially more words per battery life, making them more practical for real-time, low-power applications.

Among all models to which PTQ was applied, Llama1B achieves the highest W/BL, peaking at  $2.9 \times 10^4$ in Q2, demonstrating its efficiency in low-power conditions. Similarly, Gemma2B improves from $3.4 \times 10^3$ in FP16 to $1.2 \times 10^4$ in Q2, reinforcing the energy efficiency benefits of aggressive quantization. Llama3B follows a similar pattern, increasing from $2.8 \times 10^3$ in FP16 to $9.0 \times 10^3$ in Q2, marking a 3.2$\times$ improvement.
Larger models exhibit even more dramatic efficiency gains, improving by multiple orders of magnitude with quantization. For instance, Llama8B experiences an extreme increase from $5.9 \times 10^1$ W/BL in FP16 to $3.6 \times 10^3$ in Q2—an astonishing $61\times$ efficiency improvement. Similarly, Phi3B sees a notable boost, rising from $1.8 \times 10^2$ W/BL in FP16 to $7.5 \times 10^3$ in Q2, reflecting a more than $40\times$ gain in power efficiency. This substantial increase highlights that lower-bit quantization not only reduces energy consumption but also preserves model effectiveness, making large-scale models far more practical for deployment in battery-powered applications.
At the same time, BitNet0.7B\_Q1.58 continues to demonstrate outstanding efficiency, achieving the highest W/BL at $3.1 \times 10^4$—surpassing even the best-performing Q2 models of other architectures. Notably, BitNet0.7B\_Q1.58 is $1.07\times$ more energy-efficient than Llama1B\_Q2, underscoring the advantages of its optimized bitwise computations over conventional quantized transformers.

These results reinforce the critical role of quantization in optimizing energy efficiency, particularly for scaling large models to power-limited environments. While smaller models already perform well at low bit-widths, larger models see the most substantial improvements, making quantization an essential technique for real-world deployment.

\begin{figure}[]
    \centering
    \includegraphics[width=1.0\linewidth]{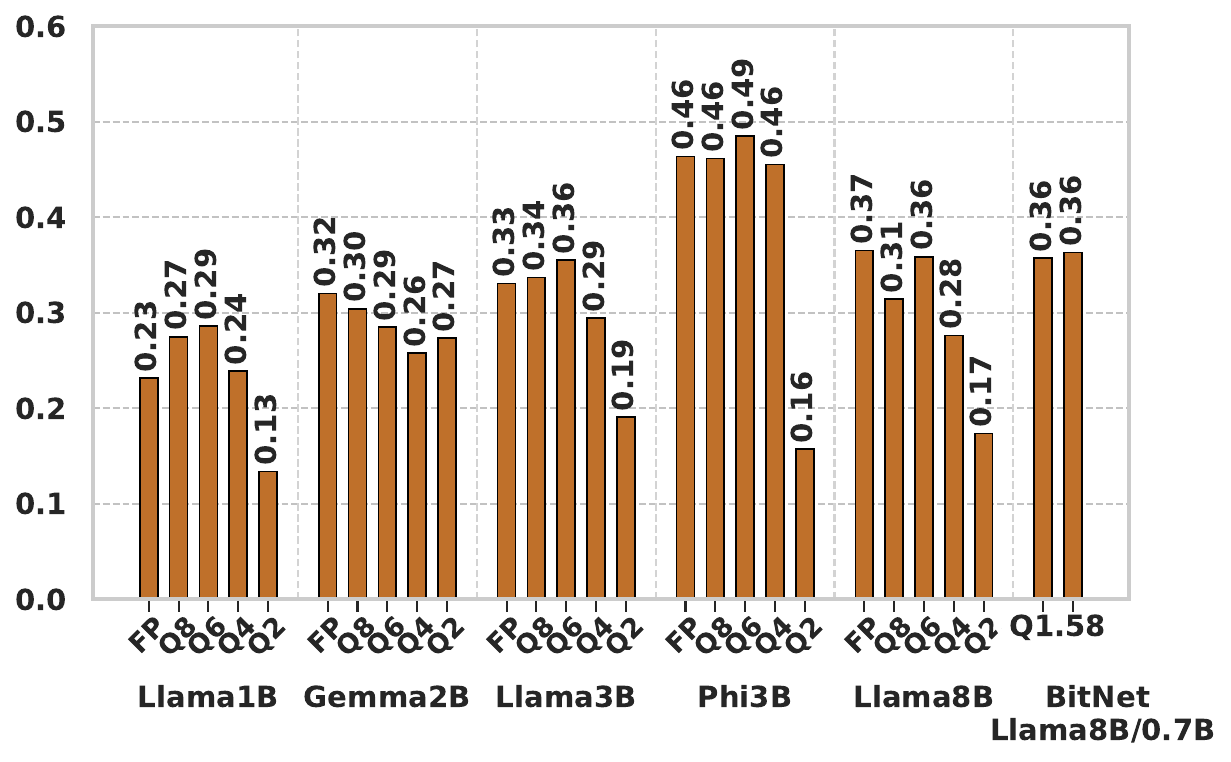}
    \caption{ NUBIA score for different LLMs and precision levels.}
    \label{fig:nubia_score}
\end{figure}

\subsection{NUBIA Score}
Figure \ref{fig:nubia_score} compares NUBIA scores across various LLMs under different precision settings, measuring semantic similarity between model-generated responses and ground-truth references. Higher NUBIA scores indicate better contextual coherence, making this a crucial metric for evaluating quantization effects. 

The results indicate that aggressive quantization generally reduces accuracy, with lower-bit precision leading to declining NUBIA scores. However, the degree of accuracy loss varies across different architectures, demonstrating that some models are more resilient to quantization than others.

Phi3B demonstrates remarkable stability across different quantization levels, maintaining NUBIA scores between 0.46 and 0.49 in FP16, Q8, Q6, and Q4. Its highest accuracy is observed at Q6 (0.49), slightly surpassing the scores of FP16, Q8, and Q4 (0.46). This resilience to precision loss suggests that Phi3B's architecture is inherently robust to quantization, likely benefiting from better weight distribution and efficient handling of lower-precision arithmetic. However, at Q2, performance drops sharply to 0.16, indicating that extreme quantization begins to degrade response quality significantly.

Llama3B improves slightly at Q8 (0.34) and peaks at Q6 (0.36) before degrading at Q4 (0.29) and Q2 (0.19). While its overall NUBIA scores are lower than Phi3B, the trend is identical, reinforcing the idea that both models benefit from moderate quantization but suffer under extreme bit-width reductions.

In Llama1B, the FP16 configuration starts at 0.23, with moderate quantization improving accuracy to 0.27 in Q8 and 0.29 in Q6. This suggests that quantization might act as a regularization mechanism, reducing overfitting and encouraging the model to rely on more generalizable linguistic patterns—a behavior similar to weight decay or dropout in deep learning models \cite{liang2021pruning}. 
However, as quantization becomes more aggressive, accuracy declines to 0.24 in Q4 and further drops to 0.13 in Q2, indicating that excessive bit reduction leads to significant information loss, outweighing any potential generalization benefits.

For Gemma2B, FP16 achieves the highest accuracy at 0.32, making it the most accurate configuration for this model. As quantization is applied, accuracy declines to 0.30 in Q8 and 0.29 in Q6. However, an interesting trend emerges at Q2, where the NUBIA score increases to 0.27, surpassing Q4 (0.26). This suggests that Q2 might introduce some compensatory effects, potentially linked to activation scaling or improved generalization under extreme compression.

Llama8B, the largest model in this study, accuracy fluctuates under different quantization levels. FP16 achieves the highest NUBIA score (0.37), outperforming most configurations except Phi3B. Moderate quantization (Q8–Q6) retains performance relatively well, but Q2 leads to a sharp drop to 0.17. This suggests that while Llama8B can tolerate some level of quantization, extreme bit reduction significantly impacts its ability to maintain contextual coherence. The varying performance across precision levels indicates that larger models do not always degrade linearly with quantization but instead exhibit sensitivity to specific bit-width thresholds, making them more dependent on higher numerical precision for preserving linguistic expressiveness. 

QAT, as used for Llama8B in the BitNet architecture, enables models to retain accuracy even under extreme quantization. Unlike PTQ, QAT optimizes weight representations during training, allowing BitNet0.7B\_Q1.58 to achieve a NUBIA score of 0.36, comparable to Llama8B’s FP16 performance. This demonstrates QAT’s effectiveness in preserving accuracy, showcasing its advantage over PTQ for deploying LLMs in resource-limited settings.

\begin{figure}[]
   \centering
   \includegraphics[width=0.98\linewidth]{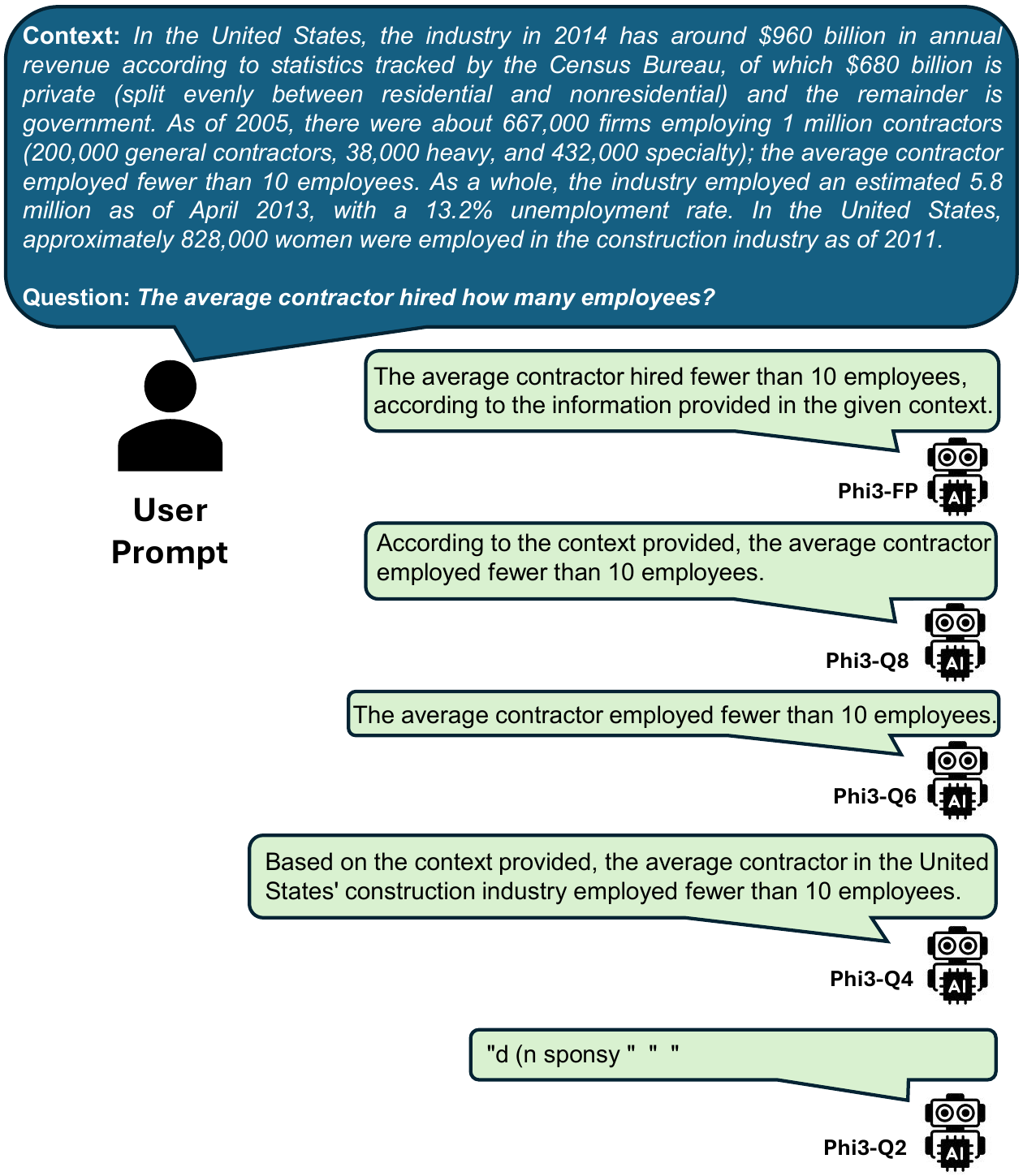}
   \caption{LLM response of Phi3B for different quantization levels.}
   \label{fig:LLM_response}
\end{figure}

To further illustrate the impact of quantization on LLM responses, Figure \ref{fig:LLM_response} provides examples of Phi3B model's generated outputs across different precision levels. While responses remain accurate and coherent from FP16 to Q4, at Q2, the model generates nonsensical output, indicating that extreme quantization disrupts semantic consistency.



\subsection{Putting Everything Together}
Our findings reveal a critical balance between accuracy, latency, and energy efficiency in deploying LLMs on edge devices. Lower-bit quantization emerges as a powerful optimization strategy, significantly enhancing real-time performance while maintaining competitive accuracy. However, the extent of this benefit varies across architectures, emphasizing the need for model-aware quantization strategies.

\begin{figure}
    \centering
    \includegraphics[width=1.0\linewidth]{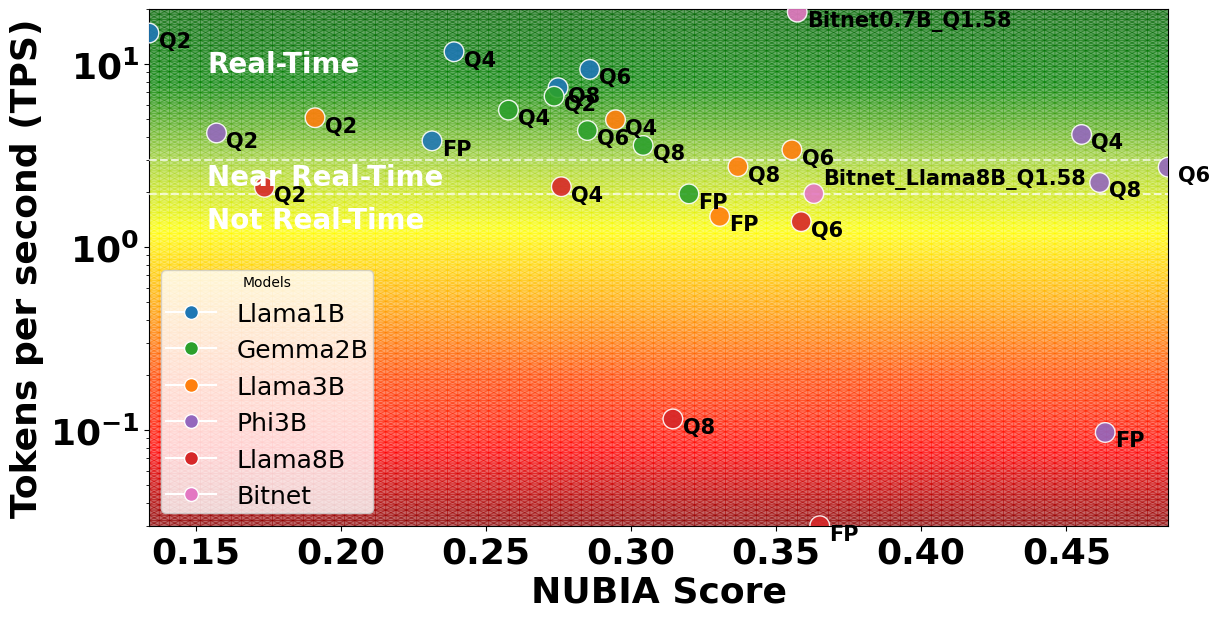}
    \caption{NUBIA Score vs. Tokens per Second (TPS) across different LLMs and quantization levels.}
    \label{fig:Accuracy_over_latency}
\end{figure}

Figure \ref{fig:Accuracy_over_latency} visualizes the accuracy-latency trade-offs, demonstrating that Phi3B and BitNet models exhibit remarkable resilience to quantization, maintaining high NUBIA scores even at lower bit-widths. In contrast, larger models like Llama8B show substantial degradation, particularly at extreme quantization levels. This underscores the fundamental role of architectural design in mitigating accuracy loss under aggressive compression.

BitNet models, with ternary Q1.58 quantization, achieve real-time inference with minimal accuracy loss, surpassing even the best PTQ-quantized transformers. Meanwhile, PTQ-optimized models such as Llama1B, Gemma2B, and Phi3B demonstrate that PTQ can serve as a viable alternative to QAT, offering both efficiency and practical deployment feasibility. 

These findings emphasize the importance of balancing accuracy, latency, and power efficiency for LLMs on edge devices, paving the way for efficient conversational AI, mobile assistants, and embedded NLP applications.

\section{Conclusion}
This study demonstrates that quantization is a highly effective strategy for optimizing LLMs for energy-efficient edge deployment. By applying PTQ across multiple bit-widths (Q8, Q6, Q4, and Q2), we significantly enhance energy efficiency, measured through TPJ and W/BL, while maintaining competitive inference speed.
The results confirm that full-precision models (FP16) exhibit the lowest energy efficiency, making them impractical for power-constrained environments. As quantization levels decrease, models generate significantly more tokens per unit of energy, making them more suitable for real-time, low-power applications. Lower-bit quantization (Q2 and Q4) achieves substantial efficiency improvements, with 4$\times$ to 60$\times$ gains, depending on the model. Among all tested configurations, BitNet0.7B Q1.58 achieves the highest energy efficiency, surpassing even the best-performing Q2 models.


Future work explore hybrid quantization, structured pruning, and hardware-aware optimizations to further enhance efficiency while maintaining low latency and high response quality.

\section*{Acknowledgments}
This work is supported in part by the National Science Foundation (NSF) under grant number 2340249.

{
    \small
    \bibliographystyle{ieeenat_fullname}
    \bibliography{main}
}


\end{document}